\documentclass[11pt,a4paper]{article}
\usepackage[T5,T2A,T1]{fontenc}
\usepackage[utf8]{inputenc}
\usepackage[hyperref]{acl2019}
\usepackage{times}
\usepackage{latexsym}
\usepackage{microtype}
\usepackage{graphicx}
\usepackage{url}
\usepackage{amsmath}
\usepackage{amsfonts}
\usepackage{dirtytalk}
\usepackage{bbm}
\usepackage{xcolor}
\usepackage{pifont}

\usepackage{tabularx,booktabs,ragged2e}

\definecolor{niceGreen}{rgb}{46,139,87}

\aclfinalcopy 
\DeclareMathOperator*{\argmax}{\arg\max}   

\title{\textit{Hello, It's GPT-2 - How Can I Help You?} \\ Towards the Use of Pretrained Language Models \\ for Task-Oriented Dialogue Systems}

\author{Pawe{\l} Budzianowski$^{1,}$$^{2,}$$^3$  \and  Ivan Vuli\'{c}$^{2,}$$^3$\\
   ${}^1$Engineering Department,   Cambridge University, UK \\
  ${}^2$Language Technology Lab,   Cambridge University, UK\\
  ${}^3$PolyAI Limited, London, UK \\
  \texttt{pfb30@cam.ac.uk}, \texttt{iv250@cam.ac.uk}\\
  }
\date{}

\begin{document}
\maketitle
\begin{abstract}
Data scarcity is a long-standing and crucial challenge that hinders quick development of task-oriented dialogue systems across multiple domains: task-oriented dialogue models are expected to learn grammar, syntax, dialogue reasoning, decision making, and language generation from absurdly small amounts of task-specific data. In this paper, we demonstrate that recent progress in language modeling pre-training and transfer learning shows promise to overcome this problem. We propose a task-oriented dialogue model that operates solely on text input: it effectively bypasses explicit policy and language generation modules. Building on top of the TransferTransfo framework \cite{wolf2019transfertransfo} and generative model pre-training \cite{radford2019language}, we validate the approach on complex multi-domain task-oriented dialogues from the MultiWOZ dataset. Our automatic and human evaluations show that the proposed model is on par with a strong task-specific neural baseline. In the long run, our approach holds promise to mitigate the data scarcity problem, and to support the construction of more engaging and more eloquent task-oriented conversational agents.
\end{abstract}

\section{Introduction}
Statistical conversational systems can be roughly clustered into two main categories: 1) task-oriented modular systems and 2) open-domain chit-chat neural models. The former typically consist of independently trained constituent modules such as language understanding, dialogue management, and response generation. The main goal of such systems is to provide meaningful system responses which are invaluable in building conversational agents of practical value for restricted domains and tasks. However, data collection and annotation for such systems is complex, time-intensive, expensive, and not easily transferable \cite{young2013pomdp}. On the other hand, open-domain conversational bots \cite{li2017adversarial, serban2017hierarchical} can leverage large amounts of freely available unannotated data \cite{ritter2010unsupervised,henderson2019repository}. Large corpora allow for training end-to-end neural models, which typically rely on sequence-to-sequence architectures \cite{sutskever2014sequence}. Although highly data-driven, such systems are prone to producing unreliable and meaningless responses, which impedes their deployment in the actual conversational applications \cite{li2017adversarial}. 

Due to the unresolved issues with the end-to-end architectures, the focus has been extended to retrieval-based models. Here, the massive datasets can be leveraged to aid task-specific applications \cite{kannan2016smart,henderson2017efficient,henderson2019training}. The retrieval systems allow for the full control over system responses, but the behaviour of the system is often highly predictable. It also depends on the pre-existing set of responses, and the coverage is typically insufficient for a multitude of domains and tasks. However, recent progress in training high-capacity language models (e.g., GPT, GPT-2)  \cite{radford2018improving, radford2019language} on large datasets reopens the question of whether such generative models can support task-oriented dialogue applications. Recently, \citet{wolf2019transfertransfo} and \citet{golovanov2019large} showed that the GPT model, once fine-tuned, can be useful in the domain of personal conversations. In short, their approach led to substantial improvements on the Persona-Chat dataset \cite{zhang2018personalizing}, showcasing the potential of exploiting large pretrained generative models in the conversational domain.\footnote{E.g., TransferTransfo \cite{wolf2019transfertransfo} yields gains in all crucial dialogue evaluation measures such as fluency, consistency and engagingness on the Persona-Chat dataset.}

In this paper, we demonstrate that large generative models pretrained on large general-domain corpora can support \textit{task-oriented dialogue applications}. We first discuss how to combine a set of diverse components such as word tokenization, multi-task learning, and probabilistic sampling to support task-oriented applications. We then show how to adapt the task-oriented dialogue framework to operate entirely on text input, effectively bypassing an explicit dialogue management module and a domain-specific natural language generation module. The proposed model operates entirely in the sequence-to-sequence fashion, consuming only simple text as input. The entire dialogue context, which includes the belief state, the database state and previous turns, is provided to the decoder as raw text. The proposed model follows the recently proposed TransferTransfo framework \cite{wolf2019transfertransfo}, and relies on pretrained models from the GPT family \cite{radford2018improving,radford2019language}.

\begin{figure}[t!]
  \includegraphics[width=0.95\linewidth]{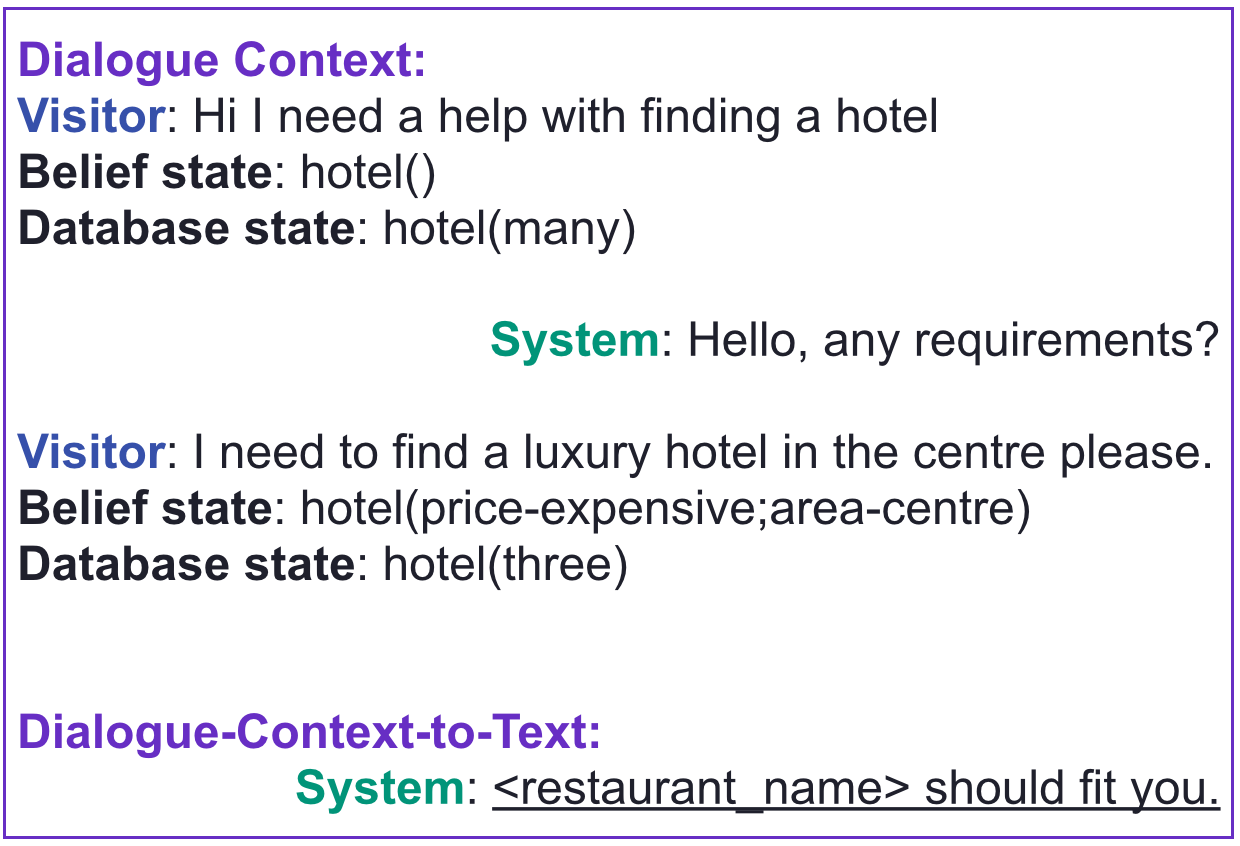}
  \vspace{-1.5mm}
  \caption{Dialogue-context-to-text task.}
  \label{fig:dialogue}
  \vspace{-3.5mm}
\end{figure}

Our results in the standard Dialogue-Context-to-Text task (see Figure~\ref{fig:dialogue}) on the multi-domain MultiWOZ dataset \cite{budzianowski2018multiwoz} suggest that our GPT-based task-oriented dialogue model learns to generate and understand domain-specific tokens, which in turn leads to a seamless adaptation to particular focused domains. While automatic evaluation indicates that our framework still falls slightly short of a strong task-specific neural baseline, it also hints at the main advantage of our framework: it is widely portable and easily adaptable to a large number of domains, bypassing the intricate modular design only at a small cost in performance. Furthermore, user-centered evaluations suggest that there is no significant difference between the two models. 

\section{From Unsupervised Pretraining to Dialogue Modeling}
Task-oriented dialogue modeling requires substantial amounts of domain-specific manually labeled data. A natural question to ask is: Can we leverage transfer learning through generative pretraining on large unlabelled corpora to enable task-oriented dialogue modeling. In this work, we rely on the standard language modeling (LM) pretraining, where the task is to predict the next word given the preceding word sequence \cite{bengio2003neural}. The objective maximizes the likelihood over the word sequence $S= \{w_1, ..., w_{|S|} \}$:
  \vspace{-1em}
\begin{equation}
    \label{eq:loss1}
    \mathcal{L}_1(S) = \sum_{i=1}^{|S|} \log P(w_i | w_0, w_1, ..., w_{i-1}).
\end{equation}

\noindent Transfer learning based on such LM pretraining combined with the Transformer decoder model  \cite{vaswani2017attention} resulted in significant progress across many downstream tasks \cite{rei2017semi,howard2018universal,radford2018improving, radford2019language}.

\subsection{TransferTransfo Framework}
\label{ss:tt}
\begin{figure*}[t!]
  \includegraphics[width=\linewidth]{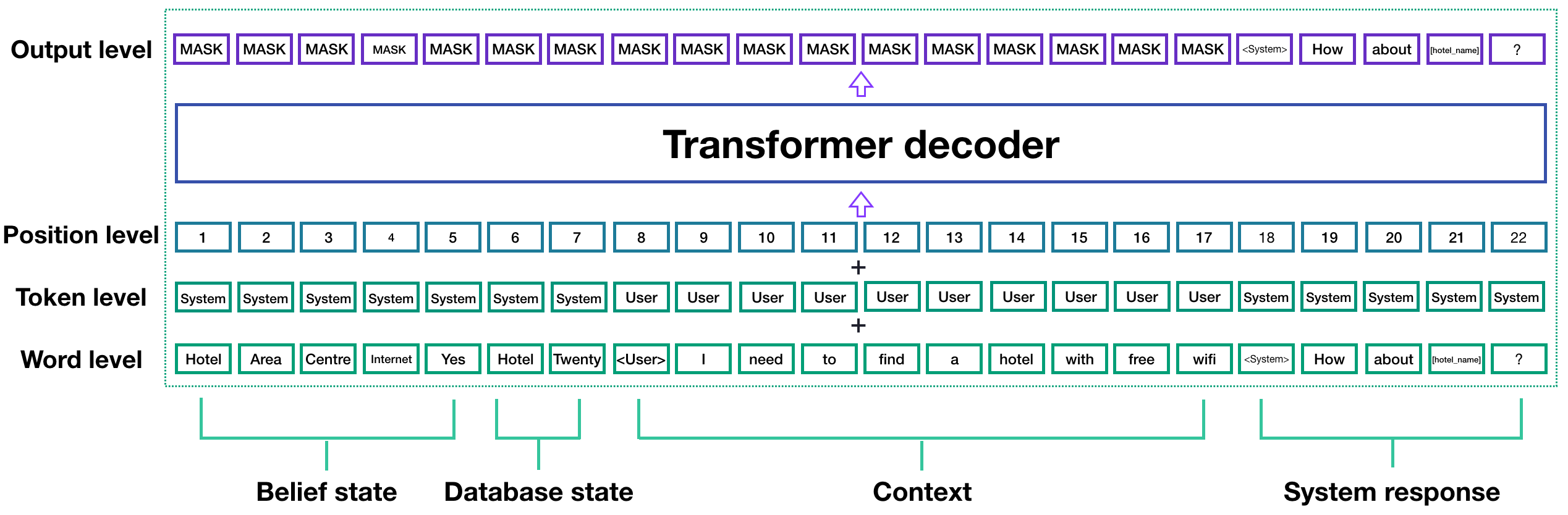}
  \caption{The framework for modeling task-oriented conversations based on a pretrained GPT model which uses only unstructured simple text as input. The context, belief state, and database state are joined together without explicit standalone dialogue policy and generation modules. The token-level (i.e., dialogue-state) embeddings are learned following \newcite{wolf2019transfertransfo}.}
  \label{fig:model}
    \vspace{-1em}
\end{figure*}

\citet{golovanov2019large} and \citet{wolf2019transfertransfo} achieved a first successful transfer of a generative pretrained GPT model to an open-domain dialogue task. The pretrained GPT model is fine-tuned in a multi-task learning fashion following the original work \cite{radford2018improving}. The LM objective from Eq.~\eqref{eq:loss1} is combined with the next utterance classification task:
\begin{equation}
\label{eq:loss2}
    p(c,a) = \text{softmax} (h_l * W_h).
\end{equation}
$c$ and $a$ represent the context of the conversation ($c$) and a proposed answer ($a$), $h_l$ is the last hidden state of the transformer decoder, and $W_h$ is learnt during the fine-tuning phase. The model significantly improves upon previous baselines over all automatic dialogue evaluation metrics as well as in evaluation with human subjects when evaluated on the Persona-Chat dataset \cite{zhang2018personalizing}.

The GPT input consists of token embeddings and positional embeddings. In order to move from a single-speaker setting to a setting with two interlocutors, \citet{wolf2019transfertransfo} introduced \emph{dialogue-state} embeddings. These embeddings inform the model whether the current token comes from an utterance of the first speaker or an utterance of the second speaker. The dialogue-state embeddings are learned during the fine-tuning phase.

\section{Domain Transfer for (Task-Oriented) Dialogue Modeling}
We now briefly discuss several advances in modeling of natural language that facilitate applicability of pretrained generative models in task-oriented dialogue modeling. To the best of our knowledge, this work is first to combine these existing components to enable task-oriented dialogue modeling.

\subsection{Domain Adaptation and Delexicalization}
Dealing with out-of-vocabulary (OOV) words has been a long-standing challenge in dialogue modeling, e.g., it is crucial for task-oriented generation where the generated output is often \emph{delexicalized} \cite{wensclstm15}. Delexicalization replaces slot values by their corresponding (generic) slot tokens and it allows learning value-independent parameters. Recently, owing to subword-level tokenisation \cite{sennrich2016neural}, language models are now able to deal with OOVs and domain-specific vocabularies more effectively \cite{radford2018improving}.

\subsection{Simple Text-Only Input}
There have been some empirical validations recently which suggest that posing NLP tasks in the form of simple text can yield improvements with unsupervised architectures \cite{wolf2019transfertransfo,radford2019language}. For instance, in task-oriented dialogue modeling the Sequicity model \cite{lei2018sequicity} sees the classification over the belief state as a generation problem. That way, the entire dialogue model pipeline is based on the sequence-to-sequence architecture: the output from one model is the input to the subsequent recurrent model. We follow this approach by providing both the belief state and the knowledge base state in a simple text format to the generator. This significantly simplifies the paradigm of building task-oriented models: any new source of information can be simply added to as another part of the text-only input provided in ``natural language''.


\subsection{Transferring Language Generation Capabilities}
Transformer architecture shows ability to learn new (i.e., domain-specific) token embeddings in the fine-tuning phase \cite{radford2018improving,wolf2019transfertransfo}. This means that the GPT models can adapt through special tokens to particular tasks.  By providing the input representation as text with domain-specific tokens, we can use off-the-shelf architectures and adapt to the domain-specific input without the need of training new dialogue sub-modules. As mentioned in \S\ref{ss:tt}, the token level layer (Figure~\ref{fig:model}) informs the transformer decoder what part of the input comes from the system side or from the user side. In our framework, we create two task-oriented specific tokens (\verb|System| and \verb|User| tokens) that are learned during fine-tuning.


\subsection{Generation Quality}
Finally, the long-standing problem of dull and repetitive response generation \cite{li2017adversarial} has  been in the focus of recent work \cite{kulikov2018importance,holtzman2019curious}. Owing to new sampling strategies, generative models are now able to create longer and more coherent sequence outputs. This has been validated also for open-domain dialogue modeling \cite{wolf2019transfertransfo,golovanov2019large}. We experiment with standard decoding strategies as well as with the recently proposed \emph{nucleus} sampling procedure \cite{holtzman2019curious}. A standard \textit{greedy sampling} strategy chooses the most probable word as :
$$\argmax_{w_i}  = \log P(w_i | w_0, w_1,..., w_{i-1}).$$
On the other hand, nucleus sampling is restricted only to words from the $p$-th percentile of the distribution during generation. The probabilities of words for which the cumulative sum exceeds the percentile are rescaled and the sequence is sampled from this subset. We probe the ability of such large pretrained models to generate more varied and semantically richer responses relying on nucleus sampling in lieu of greedy sampling without hurting the actual performance.

\section{Fine-Tuning GPT on MultiWOZ}
\label{sec:model}
To evaluate the ability of transferring the GPT generation capability to constrained/focused dialogue tasks and domains, we rely on the multi-domain MultiWOZ dataset \cite{budzianowski2018multiwoz}.
MultiWOZ consists of $7$ domains and $10,438$ dialogues and it is substantially larger than previous available datasets \cite{wen2016network,ElAsri:2017sigdial}. 
The conversations are natural as they were gathered through human-human interactions. However, the dialogues are based on domain-specific vocabulary such as booking IDs or telephone numbers that need to be delexicalized as they are entirely database-dependent. 

\paragraph{Natural Language as (the Only) Input.} 
GPT operates solely on the text input. This is in opposition to the standard task-oriented dialogue architectures \cite{wen2016network,zhao2017learning} where the belief state and the database state are encoded in a numerical form. For example, the database state is typically defined as $n$-bin encodings representing a number of available entities at the current state of the conversation \cite{wen2016network}. Therefore, we transform the belief state and the knowledge base representation to a simple text representation. The belief state takes the following form:
{
\begin{verbatim}
Domain1 Slot1 Value1 Slot2 Value2 
Domain2 Slot1 ...
\end{verbatim}}
\noindent and the database representation is provided as:
{
\begin{verbatim}
Domain1 # of entities
Domain2 # of entities ...
\end{verbatim}} 

\noindent This is also similar in spirit to the Sequicity architecture \cite{lei2018sequicity} where the second recurrent model takes as input the belief state in the natural language (i.e., simple text-only) form. In this work, we also transform the knowledge base state to a similar natural language format. These two pieces of information are then concatenated with the history of the conversation forming the full dialogue context, see Figure~\ref{fig:model}. Following \citet{wolf2019transfertransfo}, we add new token embeddings for two parties involved in the conversation to inform the attention layers what part of the context comes from the user, and what part is related to the system. Figure~\ref{fig:model} presents the final architecture.

\paragraph{Training Details.}
We use the open-source implementation of the GPT architecture that provides both GPT and GPT-2 fine-tunable checkpoints.\footnote{\url{https://github.com/huggingface/transfer-learning-conv-ai}} Following previous work \cite{radford2018improving,wolf2019transfertransfo}, we set the weight on the language model loss to be two times higher than the one for the response prediction. The parameters for the batch size ($24$), learning rate (1e-5) and the number of candidates per sequence ($2$) were chosen based on the grid search. 
\footnote{We searched over the following values: learning rates $\in$ \{1-e4, 1-e5, 5-e6, 1-e6\}, batch sizes  $\in \{8, 12, 16, 20, 24\}$ and candidate set sizes $\in \{1, 2, 4,6\}$.} 

\section{Results and Analysis}
Following prior work \cite{budzianowski2018multiwoz,zhao2019rethinking,chen2019semantically}, our evaluation task is the dialogue-context-to-text generation task (see Figure~\ref{fig:dialogue}). Given a dialogue history, the oracle belief state and the database state, the model needs to output the adequate response. By relying on the oracle belief state, prior work has bypassed the possible errors originating from natural language understanding \cite{budzianowski2018multiwoz}. 

The main evaluation is based on the comparison between the following two models: 1) the baseline is a neural response generation model with an oracle belief state obtained from the wizard annotations as in \cite{budzianowski2018towards}; 2) the model proposed in \S\ref{sec:model} and shown in Figure~\ref{fig:model} that works entirely with text-only format as input (see \S\ref{sec:model}). We test all three available pretrained GPT models - the original GPT model \cite{radford2018improving}. and two GPT-2 models referred to as small (GPT2) and medium (GPT2-M) \cite{radford2019language}. 

\subsection{Evaluation with Automatic Measures}
We report scores with three standard automatic evaluation measures. Two of them relate to the dialogue task completion: whether the system has provided an appropriate entity (\emph{Inform}) and then answered all requested attributes (\emph{Success} rate). Finally, fluency is measured by the BLEU score \cite{papineni2002bleu}. 
\begin{table}[!t]
{\small
\begin{tabularx}{\linewidth}{l XXll}
\toprule
& Baseline & GPT  & GPT2-S & GPT2-M\\
\cmidrule(lr){2-5}
Inform (\%)  & \textbf{76.7} & 71.53  & 66.43 &  70.96 \\
Success (\%) & \textbf{64.63} &55.36  & 55.16 & 61.36\\
BLEU (\%) & 18.05 &17.80 & 18.02& \textbf{19.05} \\
\bottomrule
\end{tabularx}}%
\caption{\label{tab:greedy} Evaluation on MultiWOZ with the greedy sampling procedure.}
  \vspace{-.7em}
\end{table}

\begin{table}[!t]
{\small
\begin{tabularx}{\linewidth}{l XX ll}
\toprule
& Baseline & GPT  & GPT2-S & GPT2-M\\
\cmidrule(lr){2-5}
Inform (\%)  & 72.57 & 70.43 & 69.3 & \textbf{73.96}\\
Success (\%) & 57.63 & 51.0 & 54.93  & \textbf{61.20} \\
BLEU (\%) & 15.75 & 15.65 & 15.64 & \textbf{16.55}\\
\bottomrule
\end{tabularx}}%
\caption{\label{tab:nucleus}Evaluation on MultiWOZ with the nucleus sampling procedure.}
  \vspace{-1em}
\end{table}

First, three versions of GPT were fine-tuned on MultiWOZ and evaluated with greedy sampling. The results are summarized in Table~\ref{tab:greedy}). They show that the baseline obtains the highest score on task-related metrics while the highest BLUE score was achieved by GPT2-M. Although the results are lower for the GPT-based methods, we note the design simplicity of the GPT-based task-oriented dialogue models. Further, the gap in performance might be partially attributed to the chosen greedy sampling procedure which puts too much focus on the properties of the original pretraining phase \cite{holtzman2019curious}.

Therefore, we also report the results with the {nucleus} sampling method in Table~\ref{tab:nucleus}. The scores confirm the importance of choosing the correct sampling method. The GPT2 models improve the score on \emph{Inform} and \emph{Success} metrics. It is worth noting the consistent drop in BLUE scores across all models. This comes from the fact that nucleus sampling allows for increased variability: this might reduce the probability of generating domain-specific tokens. 

We have also qualitatively analyzed a sample of successful dialogues. Only around $50\%$ of dialogues are successful both with the baseline and with the GPT-based models. Moreover, there are no clearly observed distinct patterns between successful dialogues for the two model types. This suggests that they might be effectively ensembled using a ranking model to evaluate the score of each response \cite{henderson2019training}. We will investigate the complementarity of the two approaches along with ensemble methods in future work.

\subsection{Human Evaluation}
In another, now user-centered experiment, the goal was to analyze the generation quality. Turkers, native speakers of English, were asked to rate their binary preference when presented with one-turn responses from the baseline, GPT, GPT2-M and the original dialogues (\textit{Target}). The turkers were required to choose what response they prefer when presented with two responses from two different models, resulting in more than $300$ scores per each model pair.

\begin{table}[!t]
\begin{center}
\begin{tabularx}{\linewidth}{l | cc | r}
\toprule
 Model 1  ~~~~~~~~~& \multicolumn{2}{c|}{vs}  & ~~~~~~~~~Model 2 \\
\midrule
GPT & 59 \% & 41\% & Baseline\\
GPT & 46 \% &  54 \% & Target\\
GPT2 & 46 \% & 54 \%  & Target\\
GPT2 & 45 \% & 55 \% & Baseline\\
Baseline & 43 \% & 57 \% & Target\\
GPT2 & 51 \% & 49 \% & GPT\\
\bottomrule
\end{tabularx}
\end{center}
\caption{\label{tab:human} Human ranking of responses between all pairs of four analyzed models and the original responses.}
  \vspace{-1.5em}
\end{table}
The results are summarized in Table~\ref{tab:human}, while some example dialogues with responses are provided in Figure~\ref{fig:generation}. As expected, the original responses are ranked higher than all neural models with the largest difference observed between the oracle and the baseline model. Although the generated output from the GPT is strongly preferred against the neural baseline, interestingly the opposite is observed with the GPT2 model. These inconclusive results call for further analyses in future work, and also show that there are no substantial differences in the quality of generated responses when comparing the strong neural baseline and the GPT-based models.

\section{Conclusion}
In this paper, we have made a first step towards leveraging large pretrained generative models for modeling task-oriented dialogue in multiple domains. The simplicity of the fine-tuning procedure where all necessary information can be encoded as simple text enables a quick adaptation to constrained domains and domain-specific vocabularies. We hope that this framework will inform and guide future research in hope of simultaneously improving and simplifying the design of task-oriented conversational systems.

\bibliography{acl2019}
\bibliographystyle{acl_natbib}

\appendix
\begin{figure*}[t!]
\centering
  \includegraphics[width=0.9\linewidth]{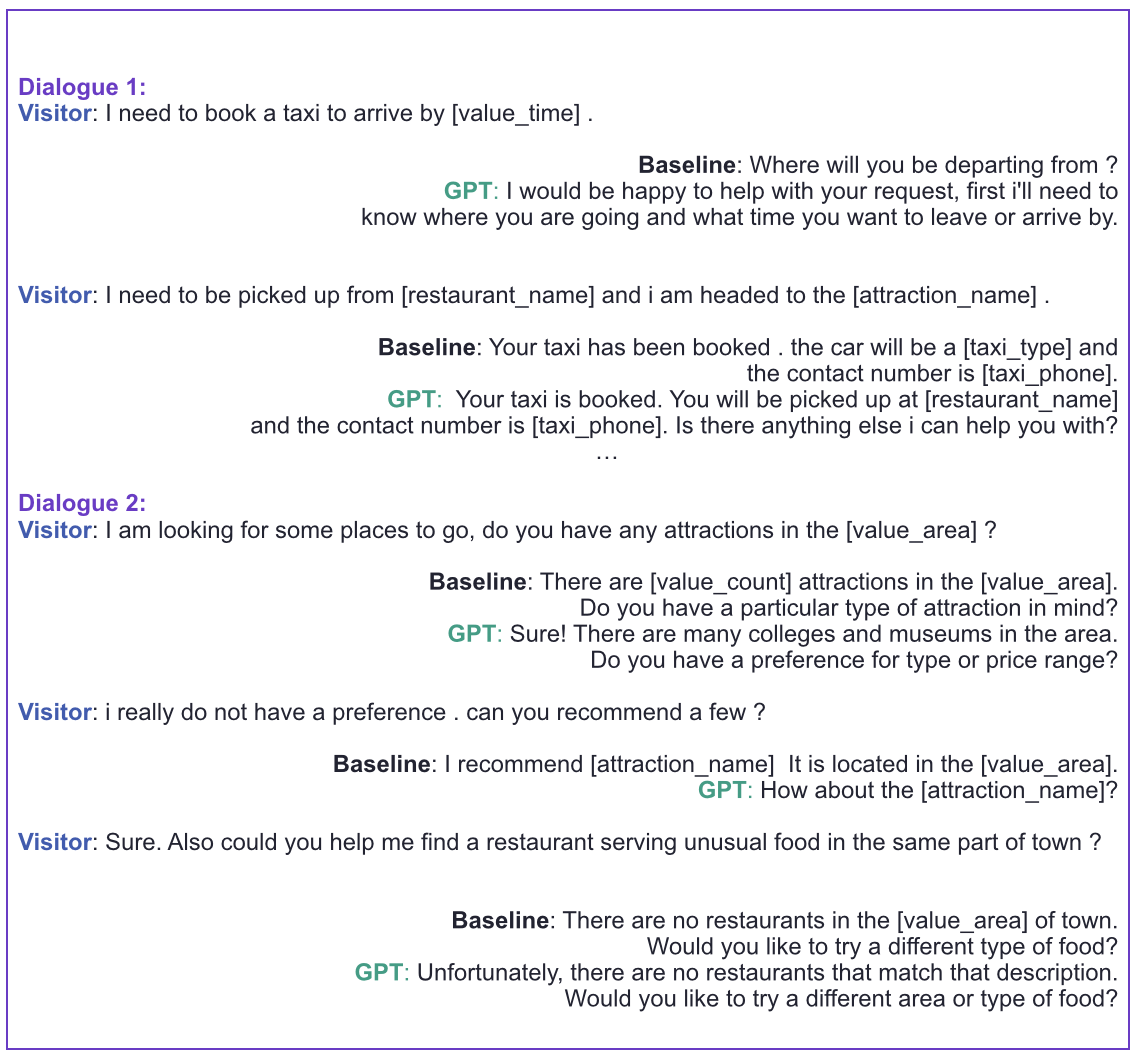}
  \caption{The comparison of generated responses from the baseline model and GPT2-M.}
  \label{fig:generation}
\end{figure*}

\end{document}